\documentclass[conference]{IEEEtran}
\IEEEoverridecommandlockouts

\def\BibTeX{{\rm B\kern-.05em{\sc i\kern-.025em b}\kern-.08em
    T\kern-.1667em\lower.7ex\hbox{E}\kern-.125emX}}
    
\usepackage{amsmath,amssymb,amsfonts}

\usepackage{bm} 
\usepackage{algorithm,algorithmic}
\usepackage{float}
\usepackage{array}
\usepackage{multirow}
\usepackage[colorlinks]{hyperref}
\usepackage{graphicx}
\usepackage{textcomp}
\usepackage [sort]{cite}
\usepackage {cleveref}
\usepackage{xcolor}
\usepackage{comment}
\usepackage{caption}

\newlength\myindent
\usepackage[compact]{titlesec}
\usepackage{dsfont}
\begin{document}

\title{Multi-Agent Deep Reinforcement Learning in Vehicular OCC}

\author{\IEEEauthorblockN{Amirul Islam, Leila Musavian, Nikolaos Thomos}
\IEEEauthorblockA{CSEE, University of Essex, UK. \\
Email: \{amirul.islam, leila.musavian, nthomos\}@essex.ac.uk}
}
\maketitle

\begin{abstract}

Optical camera communications (OCC) has emerged as a key enabling technology for the seamless operation of future autonomous vehicles. In this paper, we introduce a spectral efficiency optimization approach in vehicular OCC. Specifically, we aim at optimally adapting the modulation order and the relative speed while respecting bit error rate and latency constraints. As the optimization problem is NP-hard problem, we model the optimization problem as a Markov decision process (MDP) to enable the use of solutions that can be applied online. We then relaxed the constrained problem by employing Lagrange relaxation approach before solving it by multi-agent deep reinforcement learning (DRL). We verify the performance of our proposed scheme through extensive simulations and compare it with various variants of our approach and a random method. The evaluation shows that our system achieves significantly higher sum spectral efficiency compared to schemes under comparison.
\end{abstract}


\section{Introduction}
Autonomous vehicles are driving the revolution in future smart cities and are considered as the leading transformative technologies in intelligent transportation systems (ITS). To cope with the current ever-growing and complex nature of vehicular networks, data sharing on the road involves continuously increasing amounts of data and thus incurring enormous network overhead \cite{papadimitratos2009vehicular}. This puts tremendous pressure on the overused radio frequency (RF) spectrum that is already congested and saturated. On the contrary, the recent advancements and potential advantages of optical camera communication (OCC) over RF-based communication systems, such as license-free unlimited spectrum, lower implementation cost, and enhanced security, have rendered this technology to be an essential alternative for ITS \cite{takai2014optical}, \cite{yamazato2014image}. OCC uses light-emitting diodes (LEDs) as transmitters and cameras as receivers.

One of the main challenges of vehicular networks is that they are highly dynamic and require processing of huge amounts of data. The effectiveness of the ITS depends on supporting vehicle-to-vehicle (V2V) communication within the shortest time and lowest error. Recently, several technologies have been explored for ITS, which target delay minimization \cite{ashraf2017towards}, reliability maximization \cite{sun2014d2d} using traditional distributed method to solve the underlying optimization problems. In \cite{ashraf2017towards}, the transmission power of the vehicular network is minimized by grouping vehicles into clusters and defining reliability as queuing delay violation probability. In \cite{sun2014d2d}, a joint resource allocation and power control algorithm is proposed to maximize the communication rate considering latency and reliability constraints. However, meeting the required reliability target and at the same time respecting stringent time constraints makes the V2V communication more challenging. In particular, these problems are difficult to solve following the traditional distributed methods because of their complexity and the entailed time needed. Notably when it involves decision-making in controlling different parameters, e.g., speed, distance and modulation schemes.

Reinforcement learning (RL) can serve as an effective alternative solution to overcome the complexity of such problems \cite{sutton1998introduction}. Specifically, RL offers decentralized decision-making when the centralized decision is impossible to make. In this paper, we adopt RL, and hence, we first model the studied problem as a Markov Decision Process (MDP). Methods like value iteration that is commonly proposed to solve the MDP, require knowing the state transition probabilities beforehand making it difficult to evaluate the optimal policy. These complexities are overcome through using Q-Learning \cite{sutton1998introduction}. However, Q-Learning is characterized by slow convergence rate, and hence, is inappropriate for solving large-scale problems as the ones we study here. To address this limitation of the Q-Learning algorithm, we use deep RL (DRL) \cite{li2017deep}.

Despite overcoming the problem of Q-Learning, DRL faces difficulties in solving large-scale V2V networks in a centralized way. This introduces higher latency, which may result in increased failure rates as the vehicles may make a decision using outdated information, which eventually compromises safety. To address this issue, we utilize the concept of independent learning and multi-agent RL (MARL) \cite{tan1993multi}. In independent-learning MARL, each agent learns its policy independently and exploiting by a local observation while modelling other agents as parts of the environment dynamics.

To the best of our knowledge, DRL-based performance optimization in vehicular OCC has not been investigated in the literature. Several studies suggest the use of RL in hybrid RF and photodiode (PD)-based visible light communications (VLC) networks \cite{du2018context, kong2019q}. The authors in \cite{du2018context} apply reinforcement learning for network selection by considering the traffic type and the possibility of having learning records to improve the Q-Learning algorithm. In \cite{kong2019q}, the authors implement MARL to develop online power allocation. These works are interesting; however, they consider a centralized DRL scheme and ignore the inherent latency and reliability requirements. Moreover, they study PD-based receiver, which faces interference problems when dealing with multiple vehicles. OCC
overcomes interference problems as it can spatially separate and process different transmitter sources independently on its image plane, which has millions of pixels, and this provides freedom to handle multiple users.

In this paper, we propose independent and multi-agent DRL-based spectral efficiency maximization scheme in vehicular OCC. We maximize the spectral efficiency by adapting the modulation order from a chosen set of available modulation schemes and also by changing the relative speed of the agent (vehicle) while satisfying bit error rate (BER) and latency constraints. To the best of our knowledge, we introduce DRL for the first time in vehicular OCC for optimizing the spectral efficiency. The major contributions of this paper are summarized as follows:
\begin{itemize}
\item We formulate spectral efficiency maximization problem subject to BER, latency and a small set of modulation orders in vehicular OCC. The optimization function is a non-deterministic polynomial-time (NP) hard problem leading to a difficult search for the optimal solution. Hence, we first model the problem as an MDP;
\item We relax the constrained maximization problem by converting it to an unconstrained problem using the Lagrangian relaxation method and then solve it using deep Q-Learning;
\item We evaluate the proposed DRL-based optimization scheme and compare it with variants of the proposed scheme and a random scheme. The results show that it can effectively learn how to maximize the spectral efficiency while meeting the constraints and our solution outperforms significantly the other schemes.
\end{itemize}

\section{System Model and Problem Formulation}
\label{SM_PF}

\subsection{OCC System Model}

\begin{figure} [!t]
\centering
\setlength{\belowcaptionskip}{-10pt}
\includegraphics[width = 0.4\textwidth]{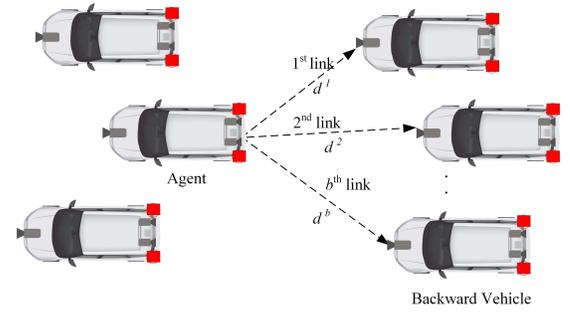}
\caption{Proposed system model for vehicular optical camera communication.}
\label{system_model}
\end{figure}

We consider a vehicular OCC system model as shown in Fig. \ref{system_model}, where each vehicle is an individual agent. Each vehicle has a transmitting unit at the back consisting of LEDs back-lights and a vision camera set and a receiving unit at the front having a high-speed camera (1000 frame per second (fps)). The camera at the back measures the backward distance using a stereo-vision camera. We consider $B$ to be the number of V2V links in the back of each vehicle and $\mathcal{B} = \{1, 2, \cdots B\}$ the set of V2V links. We express the distance of the agent (vehicle) with the backward vehicles as $d^{b}$, where $b\in \mathcal{B}$ is the index of the V2V link.{\footnote{In this paper, $\mathcal{B}$ denotes the set of V2V links.}} 

We introduce an adaptive modulation scheme of M-ary quadrature amplitude modulation (M-QAM) as it can offer low BER and improved spectral efficiency \cite{luo2015experimental}. Similar to \cite{terres2015multi}, we use time division multiple access (TDMA) in our system to transmit at different modulation orders for different backward vehicles. In TDMA, each link of the vehicle transmits at a specific time only. Hence, the spectral efficiency is divided by the number of available users, $B$, at the back.

To ensure the vehicles are free from obstruction and that can continuously communicate with each other, we assume that the transmitter and camera receiver has an uninterrupted line-of-sight link between them. The channel gain, $H^{b}_{t} $, of link $b$ in time $t$ is expressed similarly to \cite{islam2019performance} as:
\begin{align}
H^{b}_{t}  =\begin{cases}
\frac{ (m+1)A}{ 2\pi (d^{b}_{t} )^2}\; \cos^{m}(\phi) \;T_s(\theta)\; g \;  \cos (\theta) , & 0\leq  \theta \leq \theta _l\\ 
0,  & \theta > \theta _l
\end{cases}
\label{Hf}
\end{align}
where $m$ is the the order of the Lambertian radiation pattern, which is derived from LED semi-angle at half luminance, $\Phi{_{1/2}}$, as $m = \frac{-\ln (2)}{\ln (\cos (\Phi{_{1/2}}))}$. $A$ is area of the entrance pupil of the camera lens, $d^{b}_{t}$ is the agents' distance with the backward vehicles at time $t$, $\phi$ is the angle of irradiance with respect to the emitter, $T_s(\theta)$ corresponds to the transmission efficiency of the optical filter, $g$ is the gain of the lens, $\theta$ is the angle of incidence (AoI) with respect to the receiver axis, and $\theta _l$ denotes the FoV of the image sensor lens. An ideal lens has a gain: $g = {n^2}/{\sin^2(\theta _l)}$, where $n$ is the internal refractive index of the lens.

\subsection{Performance Parameter Definition}
\label{pa}
In this subsection, we specify the performance defining metrics of OCC in terms of signal-to-noise ratio (SNR), the achievable rate, and the observed transmission latency. First, we express SNR to define the communication link quality of the signal transmission. In particular, the received SNR, $\gamma ^{b}_{t} $, of the link $b$ in time $t$ for a single LED-camera communication is expressed  similarly to \cite{ashok2010challenge} as:
\begin{align}
\gamma ^{b}_{t}  = \frac{\left( \rho P_{\text{r}, t}^{b}\right)^2}{\sigma ^{b}_{t}} = \frac{\left( \rho H^{b}_{t} P \right)^2}{\sigma ^{b}_{t}} 
\label{SNR}
\end{align}
where $\rho$ is the receiver’s responsitivity, $P$ is the optical transmit power, and $\sigma ^{b}_{t}$ represents the total noise power, which is written as:
\begin{align}
\sigma^{b}_{t} = q \rho P_n A^{b}_{t} W_{\textrm{fps}} \;,
\label{PN}
\end{align}
where $q$ is electron charge, $P_n$ is the background noise power per unit area, $A^{b}_{t}$ area of the receiver for the link $b$ at time $t$, and $W_{\textrm{fps}}$ is the sampling rate of the camera in fps. We can calculate $A^{b}_{t}$ using the similar concept in \cite{horn1986robot}. To remove the effect of quantization in the received signal, measurements are made at the square grid of points. Therefore, the LED will occupy a square area of size $l'^2 = A$ having the diameter of $l' = {fl}/{d^{b}}$ of a circle since LEDs will form a circular shape on the receiver grid, where $l$ is the diameter of a LED, and $f$ is the focal length. When the projected diameter of the image becomes smaller than the size of a pixel, we refer to this as critical distance, $d_c = fl/s$, where $s$ is the edge-length of a pixel.
Based on the above definitions and from \eqref{PN}, \eqref{SNR} can be summarized as,\footnote{For notational simplicity, we drop $t$ from the notation in the remainder of the paper unless it is necessary; hence, we will adopt $\gamma^{b}$ instead of $\gamma^{b}_{t}$ and so on. Also, it is clear from the context that distance is our working variable.}
\begin{align}
\gamma ^{b} &= \begin{cases} \frac{\rho k^2 P^2}{ qP_n Wf^2l^2(d^{b})^2}; & \text{if}\; \; d^{b} < d_c\;,\\
\frac{\rho k^2 P^2}{ q P_n W s^2(d^{b})^4} ;& \text{if}\; \; d^{b} \geq d_c\;.
\end{cases}
\end{align}
where $k =  \frac{ (m+1)A}{ 2\pi}\; \cos^{m}(\phi) \;T_\text{s}(\theta)\; g \;  \cos (\theta)$.

We evaluate the BER of the optical wireless channel at the receiver using the M-QAM scheme similar to \cite{deng2017real}. Considering M-QAM, the spectral efficiency is expressed as,
  $\textrm{SE}^{b}   =\log _2 (M^{b})$,  
%
where $M^{b}$ is the available constellation points for each V2V link, $b$, e.g., $M = 4, 8, 16, \cdots$.

The channel capacity (measured in bits/sec) of the camera-based communication system is derived from the employed modulation scheme of link $b$ as has been shown in \cite{ashok2015capacity} as
\begin{align}
C^{b} =  \frac{(W_{\textrm{fps}}/3) N_{\text{LEDs}}w \varrho}{2 \;\tan \left(\frac{\theta _l}{2} \right ) \cdot d ^{b}} \cdot \log _2(M^{b}),
\label{tr_rate}
\end{align}
where $N_{\textrm{LEDs}}$ is the number of LEDs at each row of the transmitter, $w$ is the image width, and $\varrho$ is the size of LED lights in cm$^2$. Please note that, the distance $d ^{b}$ in \eqref{tr_rate} is affected by relative speed of the vehicle $v$, which also affects the position of the vehicle. The inter-vehicular distance at current time $t$ is adjusted using $d_t = d_{t-1} + v_t \cdot \Delta t$, where $d_{t-1}$ is the distance at previous time instant and $\Delta t$ is the time elapsed between time instants $t$ and $t-1$.

We consider that the end-to-end latency is dominated by the transmission latency, and we neglect the computational latency. This is because we process a small amount of data, and hence, the computational time will be short. Thus, the transmission latency of packet size, $L$, is expressed as
$
    \tau ^{b}  = {L}/{C^{b} }.
$

\subsection{Problem Formulation}
\label{PF}
Considering the proposed framework, we formulate a sum spectral efficiency optimization scheme. We aim at selecting modulation scheme from the available set of modulation orders and controlling the relative speed of the vehicle. The BER and latency are constrained so that they meet the values imposed by the system. Finally, the proposed maximization problem is formulated as:
\begin{align}
 \max _{\mathcal{M}, v} \quad & \frac{1}{B} \sum_{b = 1} ^{B}\log _2 \left( M^{b} \right)  ,
 \label{pf_sim}\\
\text {s.t.} 
\quad &  \text{BER} ^{b}  \leq \text{BER}_\text{tgt}, \; \forall b;
\label{cons_ber}\\
\quad & \tau  ^{b}  \leq \tau_ \text{max} ,\;  \forall b;
\label{cons_lat}\\
\quad & M^{b} \in \mathcal{M} , \;  \forall b;
\label{cons_M}
\end{align}
where $\mathcal{M}$ is the set of the available modulation orders, $\text{BER}_\text{tgt}$ is the maximum target BER, and $\tau_ \text{max}$ is the maximum allowed latency. Equations \eqref{cons_ber} and \eqref{cons_lat} correspond to the BER and latency constraints. The modulation scheme is chosen from a small set of available M-QAM schemes, as shown in \eqref{cons_M}.

\section{DRL-based Problem Formulation and Proposed Solution}
\label{RL_OP}

The optimization problem presented in \eqref{pf_sim} - \eqref{cons_M} is an NP-hard combinatorial problem \cite{plaisted1976some}, and thus it is challenging to find the optimal solution. The optimization problem also includes non-linear operations, such as, \eqref{pf_sim} and \eqref{cons_lat}. To solve this problem using a distributed method, each agent should choose the speed and modulation scheme separately, which makes the solution more complex and time-consuming. Recall that, in vehicular communication, we should meet low latency and required BER targets to ensure that the information is received reliably within the shortest time. Reinforcement learning is an effective way to solve such dynamic and time-varying problems as it can learn the optimal policy through interaction with the environment, and this way adjust to the environmental changes over time.

\subsection{Modeling of MDP} 
The optimization problem \eqref{pf_sim} is modelled as an MDP with a tuple ($\mathcal{S}$, $\mathcal{A}$, $p$, $r$, $\zeta$) \cite{sutton1998introduction}, where $\mathcal{S}$ is the set of all possible states; $\mathcal{A}$ denotes the set of all possible actions; $p$ denotes the transition probability $p(s_{t+1}, r_{t}|s_{t}, a_{t})$ when the agent selects an action $a_{t} \in \mathcal{A}$ and transits to a new state $s_{t+1} \in \mathcal{S}$ from the current state $s_{t} \in \mathcal{S}$; and $r$ represents the reward. While $\zeta \in [0, 1]$ is the discount factor, which gradually discounts the effect of an action to future rewards. 
The state space $\mathcal{S}$, the action space $\mathcal{A}$, and the reward function, $r$ of the considered RL framework are defined below. 

\textbf{State}: The observed states of our considered environment include: the backward distance vector, $\textbf{d}^b_{t} = (d_{t}^1, \cdots, d_{t}^B)$ and the modulation order set, $\mathcal{M}$ = $\left\lbrace \text{4, 8, 16, 32, 64} \right\rbrace$. 

\textbf{Action:} At state $s_t$, the agent takes an action $a_{t}$, by changing its relative speed, $v_t$ and selecting the modulation order from the set $\mathcal{M} $. 

\textbf{Reward:} The reward function that guides the overall learning should be consistent with the objective. Since our objective is to maximize the sum spectral efficiency, we design our reward function as a weighted sum of a reward related to the backward distance and the sum spectral efficiency \eqref{pf_sim}. First, we express the reward related to distance as follows: 
 \begin{align}
& r^{\text{d}, i}_t = \begin{cases}
- 1\times (d_\text{stop}- {d}_{t}^b) , &  {d}_{t}^b < d_\text{stop} \;,\\ 
\frac{1} {{d}_{t}^b - d_\text{stop}}, & {d}_{t}^b > d_\text{stop} \;,
\end{cases}
\label{rd}
\end{align} 
where $i$ is the index of the agent and $d_\text{stop}$ is the stopping distance \cite {AymelekgebZinchenko2014}. In our system, each vehicle will carry out the same process individually. As a result, for notational simplicity, we drop $i$ hereafter. 

Finally, considering the objective function of \eqref{pf_sim}, the overall reward, $R_{t}$, can be expressed as
\begin{align}
R_{t}=  \omega _d \;  r^{\text{d}}_t +\omega _r  \; \frac{1}{B} \sum_{b = 1} ^{B}\log _2 \left( M^{b}_{t} \right)  ,
\label{r}
\end{align}
where $\omega _d $ and $\omega _r $ are positive weights that balance distance and sum spectral efficiency rewards. The weights can be adjusted based on the system requirements.

\subsection{RL-based Problem Formulation}
The goal of RL is to maximize the expected return from the state $s_t $ by determining the optimal policy. The return, $G_{t} $, is defined as the cumulative discounted reward, and is expressed as follows:
\begin{align}
& G_{t} =  \sum _ {j=0}^{\infty} \zeta ^{j} R_{t+j+1},& 0 \leq \zeta \leq 1.
\end{align}

In summary, the objective of our proposed system is to determine the optimal policy, i.e., to select the speed and modulation order while respecting the BER and latency constraints. From the above consideration, the reward maximization problem that corresponds to the problem formulation presented in Section \ref{PF} is expressed as 
\begin{align}
 \max \quad & \mathbb{E} \left[  G_{t} \left( s_{t}, a_{t}\right) \right] ,\; \forall t
 \label{rm}\\
\text {s.t.}  
\quad &  \text{BER}^b_{t}\leq \text{BER}_\text{tgt}, \;  \forall t ;
\label{r1}\\
\quad &  \tau ^b _{t} \leq \tau_ \text{max}, \;  \forall t ;
\label{rr2}
\end{align}

\subsection{Solution of the Problem}
We can solve the constrained MDP problem  presented  in \eqref{rm}-\eqref{rr2} by converting it into an unconstrained one following Lagrange Relaxation method \cite{boyd2004convex}. Therefore,  by relaxing the BER and latency constraints, we re-express the constrained optimization problem as:
\begin{align}
c^{\bm{\lambda}, \bm{\nu}} (s_{t} , a_{t}) = \; & R_{t} \left( s_{t} , a_{t} \right)  - \sum_{b = 1} ^{B} \lambda ^b \cdot (\text{BER}^b_{t} - \text{BER}_\text{tgt})\nonumber \\ & - \sum_{b = 1} ^{B} \nu ^b \cdot (\tau ^b _{t} - \tau_ \text{max} ),
\label{cl}
\end{align}
where $\bm{\lambda} = (\lambda ^1, \lambda ^2 \cdots, \lambda ^b)$ and $\boldsymbol{\nu} = (\nu ^1, \nu ^2, \cdots, \nu ^b)$ are vectors representing the Lagrange multipliers corresponding to the constraints in \eqref{r1} and \eqref{rr2}, respectively. The optimal value of the constrained MDP problem is computed as \cite{mastronarde2012joint}:
\begin{align}
L^{\pi ^{\ast}, \bm{\lambda} ^{\ast}, \bm{\nu} ^{\ast}}_{\delta} (s) = \max _{\bm{\lambda}, \bm{\nu} \geq 0} \min _{\pi \in \phi} V^{\pi, \bm{\lambda}, \bm{\nu}} (s) - \sum_{b = 1} ^{B} \lambda ^b \delta _1 - \sum_{b = 1} ^{B}  \nu ^b \delta _2 ,
\label{clo}
\end{align}
where $\delta = \{\delta _1, \delta _2\}$, with $\delta _1 = \text{BER}_\text{tgt}$ and $\delta _2 = \tau_ \text{max}$. $\phi$ denotes the set of all possible stationary policies.

In practice, the optimal policy cannot be determined using value iteration as it requires knowledge of transition probabilities beforehand, which is not possible because of the size of the state and action space.\footnote{For notational simplicity, we drop the Lagrangian multipliers from the notation in the remainder of the paper unless it is necessary, for example, we will write $c(s_{t} , a_{t})$, $Q ^{\ast} (s_t) $, instead of $c^{\bm{\lambda}, \bm{\nu}}(s_{t} , a_{t})$, $Q ^{\ast, \bm{\lambda}, \bm{\nu}} (s_t) $, respectively.}  To solve this problem, we adopt a model-free RL approach known as tabular Q-Learning. This approach uses the $Q_t(s_{t} , a_{t})$ values for each state-action pair instead of the value function, which is a function of the state only. The Q-Learning algorithm employs the following recursive formula to update the $Q_t(s_{t} , a_{t})$ values:
\begin{align}
\begin{split}
Q_{t+1}(s_{t} , a_{t}) = & (1 - \alpha _{t})Q_t(s_{t} , a_{t})   + \alpha _{t} \left[c_{t}(s_{t} , a_{t}) \vphantom {\max _{a_{t+1}\in \mathcal{A}} } \right. \\
 &\left. + \zeta \max _{a_{t+1}\in \mathcal{A}}  Q_{t}(s_{t+1}, a_{t+1})\right],
\end{split}
\label{q_value}
\end{align}
where $ \alpha _{t}\in [0, 1] $ is a time-varying learning rate. After computing the Q-values, the optimal policy $\pi^{*}$ can be determined as
\begin{align}
\pi ^{\ast, \bm{\lambda}, \bm{\nu}} (s_t) = \arg \max_{a_t \in \mathcal{A}}\; Q^{\ast, \bm{\lambda}, \bm{\nu}}(s_{t} , a_{t}),  \forall s \in S. 
\label{opt_pi}
\end{align}

\begin{table}[!t]
\centering
\caption{Vehicular OCC modelling parameters}
\label{Notation}
\begin{tabular}{|l|l|}
\hline
\multicolumn{1}{|c|}{\textbf{Parameter, Notation}}      & \multicolumn{1}{c|}{\textbf{Value}}
\\ \hline
Angle of irradiance w.r.t. the emitter, $\phi$          & $70^o$                              \\ \hline
AoI w.r.t. the receiver axis, $\theta$                  & $60^o$               
 \\ \hline
FOV of the camera lens, $\theta _l$                     & $90^o$                               \\ \hline
Image sensor physical area,  $A$                        & $10$ cm$^2$                          \\ \hline
Optical filter Transmission efficiency, $T_s$    & 1                                    \\ \hline
Concentrator/lens gain, $g$                             & 3                                   \\ \hline
Optical transmitting  power, $P$                        & $1.2$ Watts                        
\\ \hline
Modulation scheme set, $\mathcal{M}$                                 & 4, 8, 16, 32, 64  
 \\ \hline
Camera-frame rate, $W_\text{fps}$                       & 1000 fps                                                                                                             
 \\ \hline
Number of LEDs at each row, $N_\text{LEDs}$             & 30 
\\ \hline
 Packet size, $L$                             & 5 kbits                                                                           
\\ \hline
 Size of the LED, $\varrho$                                   & 15.5 $\times$ 5.5 cm$^2$                                                               \\ \hline
 Resolution of image, $w$                               & 512 $\times$ 512  pixels                                                                     \\ \hline
\end{tabular}
\end{table}

When the state-action space is small, Q-Learning can determine the optimal policy. However, Q-Learning cannot decide the value functions or optimal policies accurately in large-scale problems within a reasonable time because of the state-action space. This problem can be solved by employing deep learning-based function approximators by means of deep neural networks. Deep Q-network (DQN) is used to train the network and learn the optimal policy. 

In order to stabilize the learning of DQN, we follow the target network approach. The DQN consists of two separate networks known as the main network that approximates the Q-function and the target network that gives the target for updating the main network. The target network is not updated after each iteration because it adjusts the main network updates to control the value estimations. If both networks are updated simultaneously, the change in the main network would be exaggerated due to the feedback loop from the target network, which results in an unstable network. 

To ensure convergence, the neural network aims to minimize the loss function, $L(\bm{\beta})$, which can be defined as
\begin{align}
L(\bm{\beta}) = \mathbb{E} \left[y_{t} - Q\left(s_{t}, a_{t}; \bm{\beta}\right)\right]^2 ,
\label{Loss}
\end{align}
where $y_{t} = c(s_{t} , a_{t}) + \zeta \max _{a_{t+1}\in \mathcal{A}} Q\left( s_{t+1},a_{t+1}; \bm{\beta} _{-}\right)$ is the target for each iteration. $\bm{\beta} $ denotes the neural network’s parameters of current iteration and $\bm{\beta}_{-}$ is the value from the previous update. Note that, $\bm{\beta}_{-}$ are held fixed when optimizing the loss function $L(\bm{\beta})$.
The optimal value of the Lagrange multipliers $\lambda^b$, $\nu ^b$  in \eqref{cl} can be learned online using a stochastic sub-gradient method as presented in \cite{salodkar2008line}.

\begin{table}[!t]
\centering
\caption{List of DRL hyper-parameters and their values}
\label{imp_para}
\begin{tabular}{|l|l|}
\hline
\multicolumn{1}{|c|}{\textbf{Parameter, Notation}}      & \multicolumn{1}{c|}{\textbf{Value}} 
 \\ \hline
Mini-batch size           & $32$  \\ \hline
Replay memory size           & $100000$  \\ \hline
Number of hidden layer (Neurons) & $1 (250)$  \\ \hline
Exploration rate, $\epsilon$              & $0.05$ \\ \hline
Discount factor, $\zeta$                  & $0.98$    \\ \hline
Activation function                     & ReLU \\ \hline
Optimizer           & RMSProp   \\ \hline 
Learning rate (used by RMSProp)           & $0.001$  \\ \hline
Gradient momentum (used by RMSProp) & $0.95$  \\ \hline
\end{tabular}
\end{table}

\section{Simulation Setup and Results}
\label{Results}

\begin{figure} [!t]
\centering
\setlength{\belowcaptionskip}{-10pt}
\includegraphics[width= .45\textwidth]{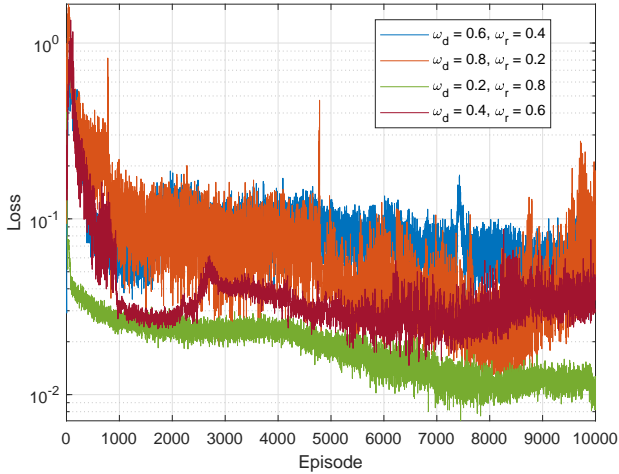}
\caption{Convergence of loss function for $\epsilon$ = 0.05 and learning rate $\alpha$ = 0.001.}
\label{Loss_ep}
\end{figure}

\subsection{Simulation Setup}
To evaluate the performance of the proposed system, we build a simulation environment upon traffic simulator Simulation of Urban Mobility (SUMO) \cite{krajzewicz2012recent}. Our simulation framework maintains the connection between SUMO and the DRL agent using Traffic Control Interface (TraCI). The DQN consists of three fully connected layers, including an input layer, a hidden layer, and an output layer. The hidden layer has 250 neurons. We use rectified linear unit (ReLU) as the activation function. We then adopt root mean square propagation (RMSProp) optimizer to minimize the loss, where we set an initial learning rate to 0.001, which varies over time. The neural network is designed using Tensorflow \cite{abadi2016tensorflow}. We implement $\epsilon$-greedy policy to balance between exploration and exploitation.

In our simulation, we train the DQN for 10000 episodes. The exploration rate, $\epsilon$, is set to 0.05. We choose a discount factor, $\zeta = 0.98$. We present the simulation parameters for the OCC system model in Table \ref{Notation}, whereas the training and testing parameters of the DRL are listed in Table \ref{imp_para}.  

We investigate the performance of the proposed multi-agent DRL-based vehicular scheme against different methods for comparison, namely, greedy, far-sighted and random scheme. In greedy method, we assume $\zeta = 0$ in \eqref{Loss}, whereas in far-sighted case, we assume $\zeta = 1$, while we keep all other parameters of the systems as reported in Table \ref{imp_para}. Finally, in the random scheme, the action is chosen randomly for all the vehicles at each time slot.

\begin{figure} [!t]
\centering
\setlength{\belowcaptionskip}{-10pt}
\includegraphics[width= 0.45\textwidth]{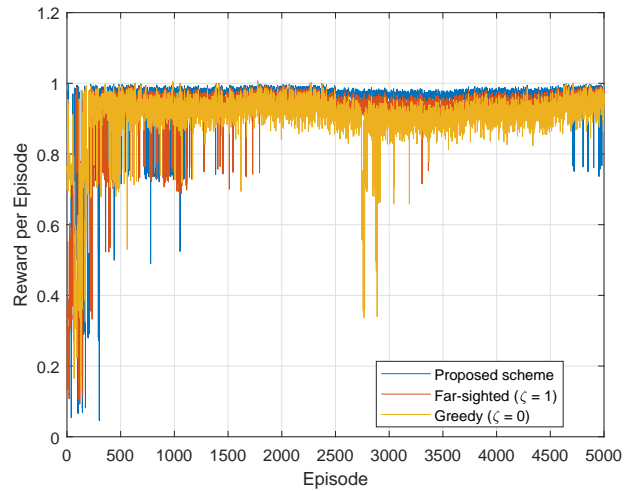}
\caption{Reward per training episode for three different approaches for $\epsilon$ = 0.05 and learning rate $\alpha$ = 0.001.}
\label{reward_ep}
\end{figure}

\subsection{Simulation Results}

First, we perform an ablation study to determine the optimal weight values for distance and spectral efficiency rewards in \eqref{r}. In doing so, we examine the proposed algorithm for different weight settings, but for simplified representation, we include four settings, such as $\omega _d = 0.2$ and $\omega _r = 0.8$, $\omega _d = 0.4$ and $\omega _r = 0.6$, $\omega _d = 0.6$ and $\omega _r = 0.4$, $\omega _d = 0.8$ and $\omega _r = 0.2$, as shown in Fig. \ref{Loss_ep}. We can see that we achieve improved loss performance when we allocate more weight value toward the spectral efficiency part. The figure points that the algorithm converges at 8000 episodes for $\omega _d = 0.2$ and $\omega _r = 0.8$. On the contrary, other weight needs more time to converge and show frequent variations in the loss and offer higher loss than $\omega _d = 0.2$ and $\omega _r = 0.8$ set. Thus, we adopt these weight values for the rest of our performance evaluation.

To analyze the convergence behaviour of the multi-agent vehicular OCC system, we demonstrate the cumulative rewards per episode for three different discount factors, i.e., the proposed scheme ($\zeta = 0.98$), greedy ($\zeta = 0$) and far-sighted ($\zeta = 1$). The results are shown in Fig. \ref{reward_ep}. From this figure, we observe that until 1500 episodes, the greedy and far-sighted approaches perform better than the proposed scheme. However, the cumulative reward for the proposed schemes improves as the training advances and eventually reaches to lower loss. We can conclude that the proposed scheme achieves higher rewards than all the schemes under consideration.

\begin{figure} [!t]
\centering
\setlength{\belowcaptionskip}{-10pt}
\includegraphics[width= 0.45\textwidth]{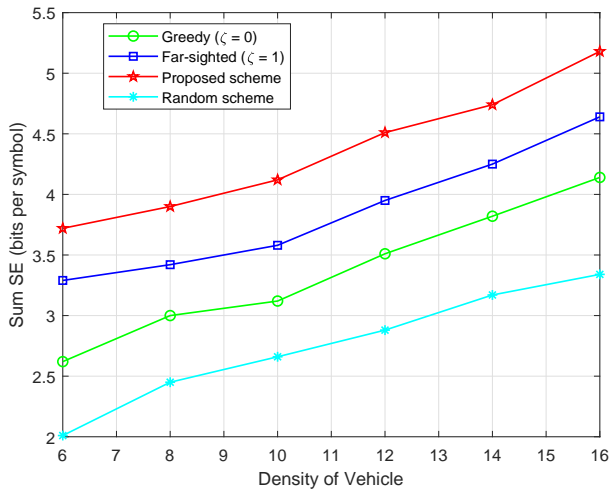}
\caption{Comparison of sum spectral efficiency with different approaches for $\epsilon$ = 0.05 and learning rate $\alpha$ = 0.001.}
\label{SE_veh}
\end{figure}

Finally, we present the maximized sum spectral efficiency versus various density of vehicles for all schemes under comparison in Fig. \ref{SE_veh}. From the figure, we see that the sum spectral efficiency increases with an increase in the density of vehicles for all the methods, and there is a significant performance gap between each scheme. We also observe that our proposed scheme can achieve a maximum of 5.2 bits per symbol, whereas the random method can achieve 3.3 bits per symbol. The results show that the proposed algorithm obtains approximately 2.3 times better rates in comparison to the random scheme, 1.25 times for far-sighted, and about 1.11 times for greedy schemes when the density of vehicles is 16. Accordingly, we can conclude that our proposed OCC system outperforms all the other schemes.

\section{Conclusion}

\label{Con}

In this paper, we present a multi-agent DRL-based spectral efficiency optimization scheme in vehicular OCC while respecting BER and latency requirements. In doing so, we optimize our system by selecting the optimal modulation order and adjusting the relative speed of each vehicle. To overcome the inherent complexity of the studied problem, we model the problem as an MDP. We then convert the constrained problem into an unconstrained problem using the Lagrangian relaxation method. Next, we solve the problem by employing deep Q-Learning to deal with the large state-action spaces we encounter. Finally, we verify the performance of our scheme through extensive simulations and compare it with various variants of our scheme. The evaluations reveal that our system achieves better sum spectral efficiency compared to the schemes under comparison.

\ifCLASSOPTIONcaptionsoff
  \newpage
\fi

\bibliographystyle{IEEEtran}
\bibliography{references}

\end{document}